\documentclass[final,1p,times]{elsarticle}

\usepackage{amssymb}
\usepackage{amsmath}

\usepackage{algorithm}
\usepackage{algpseudocode}

\usepackage{hyperref}

\usepackage{booktabs}

\usepackage[table]{xcolor}
\newcolumntype{P}[1]{>{\centering\arraybackslash}p{#1}}

\journal{Applied Energy}

\begin{document}

\begin{frontmatter}



\title{Short-Term Power Demand Forecasting for Diverse Consumer Types to Enhance Grid Planning and Synchronisation}


\author{{Asier Diaz-Iglesias\corref{cor1}$^{*1}$},{ Xabier Belaunzaran$^{1}$},{ Ane M. Florez-Tapia}$^{1}$ } 

\affiliation{organization={Vicomtech Foundation, Basque Research and Technology Alliance (BRTA)},
            city={Donostia-San Sebastián},
            postcode={20009}, 
            country={Spain}}

\begin{abstract}
Ensuring grid stability in the transition to renewable energy sources requires accurate power demand forecasting. This study addresses the need for precise forecasting by differentiating among industrial, commercial, and residential consumers through customer clusterisation, tailoring the forecasting models to capture the unique consumption patterns of each group. A feature selection process is done for each consumer type including temporal, socio-economic, and weather-related data obtained from the Copernicus Earth Observation (EO) program. A variety of AI and machine learning algorithms for Short-Term Load Forecasting (STLF) and Very Short-Term Load Forecasting (VSTLF) are explored and compared, determining the most effective approaches. With all that, the main contribution of this work are the new forecasting approaches proposed, which have demonstrated superior performance compared to simpler models, both for STLF and VSTLF, highlighting the importance of customized forecasting strategies for different consumer groups and demonstrating the impact of incorporating detailed weather data on forecasting accuracy. These advancements contribute to more reliable power demand predictions, thereby supporting grid stability.
\end{abstract}

\begin{graphicalabstract}
\begin{figure}[!ht]
    \includegraphics[width=\textwidth]{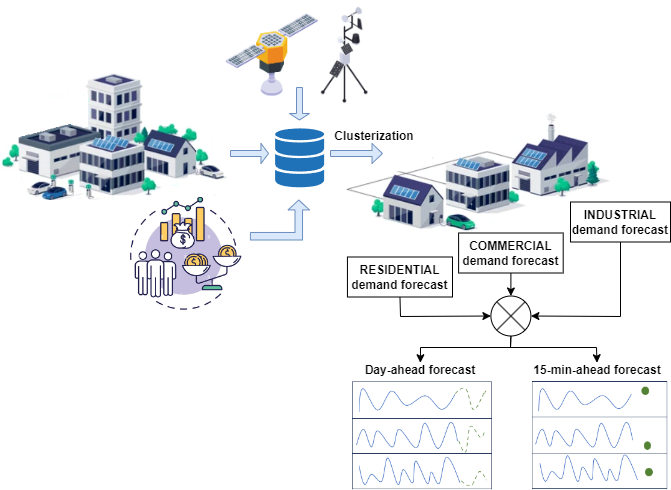}
    \label{fig:Graph_abs}
\end{figure}
\end{graphicalabstract}

\begin{highlights}
\item Machine learning is better than deep learning if data is limited for load forecasting
\item A ruled-based approach is enough to classify load profiles by consumer type
\item Weather does not affect industrial power consumption
\item Holiday data is essential when performing load forecasting for industrial consumers
\item Weather variability highly influences residential load fluctuations
\end{highlights}


\end{frontmatter}




\section{Introduction}\label{sec1}
The European Union's ambition to become a climate-neutral economy by 2050 \cite{Green_Deal} is deeply intertwined with the need for effective grid management and accurate power demand forecasting. As the EU accelerates its transition towards renewable energy sources (RES), the variability and intermittency of these resources present significant challenges for maintaining a stable and reliable power grid. Accurate demand forecasting becomes crucial in this context, as it enables grid operators to anticipate fluctuations in power demand and supply, ensuring that the balance is maintained despite the inherent variability of RES. Effective grid management, supported by precise forecasting, is vital to preventing grid disruptions like imbalances or blackouts, making it essential to the EU’s climate goals and a secure green energy transition.

Proper planning and effective applications of power demand forecasting require particular “forecasting intervals,” or “lead time”. \cite{Ahmad2020,Nti2020}. Based on the lead time, load forecasting can be categorized into four distinct types \cite{Sheng2021,hammad2020methods}:
\begin{itemize}
    \item Very-short-term load forecasting (VSTLF): applicable for real-time control with a prediction period ranging from minutes to 1 hour ahead.
    \item Short-term load forecasting (STLF): it covers forecasting within 1 hour to 7 days.
    \item Medium-term load forecasting (MTLF): it covers forecasting ranging from a few months, up to 2-3 years.
    \item Long-term load forecasting (LTLF): it covers forecasting typically ranging from 1 year to 10 years or more.
\end{itemize}

While LTLF presents the most difficult challenge in terms of forecasting \cite{Long_Term}, VSTLF and STLF are the most critical ones \cite{Holderbaum2023, Khwaja2020}. They have only become viable in recent years due to the widespread adoption of advanced metering infrastructure (AMI) and connected sensors capable of capturing electrical consumption at a high level of granularity \cite{6945384, Ribeiro2022}.

Given the substantial increase in the use of AMIs, this paper focuses on VSTLF and STLF of individual consumers located in two different Spanish regions with varying climatic conditions. The consumers are categorized into three types: industrial, commercial, and residential \cite{Toit2016}. Each consumer type has unique temporal characteristics that must be taken into account when applying time-series forecasting algorithms; e.g., while commercial and industrial consumers typically exhibit pronounced seasonality patterns due to the workweek, the load of residential consumers is usually highly variable and irregular.

In addition to the temporal characteristics, socio-economic factors and the climatic influence on each consumer need to be considered. Socio-economic factors significantly influence demand forecasting across different regions and consumer segments by shaping consumption patterns and energy usage \cite{delReal, Hu2020}. Accurate forecasting models must account for these dynamics to reflect the variability in power usage effectively.

Weather also plays a crucial part in power demand forecasting, directly influencing energy consumption patterns. Weather variables can significantly impact the need for heating, cooling, and lighting, leading to variations in demand, therefore, accurate forecasting models must account for these variables to predict energy usage effectively \cite{Perera2020}. Hence, spatially representative weather data are used to plan, design, size, construct, and manage energy systems for performance analysis and forecasting to enhance system efficiency, \cite{Amin2024}. Specifically, this article proposes leveraging the services of the Copernicus Earth Observation (EO) programme and the European Centre for Medium-Range Weather Forecasts (ECMWF), obtaining data from the ERA5  dataset \cite{era5}, which provides comprehensive, high-resolution, and globally consistent datasets. Data from national weather stations was also used, to enhance prediction accuracy.

Load forecasting models have been categorized into statistical, intelligent-computing-based, and hybrid models \cite{STLF_review}. In recent years, Artificial Intelligence (AI) methods have emerged as powerful tools for improving the accuracy of VSTLF and STLF.  These approaches include commonly used support vector machines (SVM) or artificial neural networks (ANN). More recently, emphasis has been placed on deep learning techniques such as convolutional neural networks (CNN), long short-term memory (LSTM), or transformers \cite{CNN_ref, LSTM_ref, Transformer_ref}. While deep learning has shown outstanding performance, it requires large amounts of data and careful pre-processing of the data. This paper, therefore, will examine different AI-based approaches and will elucidate a process of developing efficient and robust forecasting models for different types of customers with a focus on the 15-minute and 24-hour time horizons, to achieve the goals set in terms of accuracy.

Our study, conducted within the framework of a Horizon Europe project, offers several significant contributions. First, we have developed a simple yet accurate customer classifier that effectively distinguishes between different consumer types. Second, we present a comprehensive analysis to identify and prioritize key features for each consumer type, ensuring a deep understanding of the unique characteristics of each group. Third, we explore and compare various AI and machine learning algorithms for load forecasting to identify the most effective approaches. However, the primary contribution of our work is the introduction of novel forecasting methods tailored to each consumer type and applicable to both very short-term load forecasting (VSTLF) and short-term load forecasting (STLF). These methods outperform traditional, simpler approaches and offer distribution system operators (DSOs) valuable tools for managing the electrical grid with greater precision and efficiency.

The structure of this paper is as follows: First, a comprehensive analysis of the current state of the art is conducted. Next, the proposed methodology is outlined, followed by the presentation of the case study and a discussion of the results. Finally, the paper concludes with a discussion of future work and key conclusions.

\section{Related work}

This section presents an overview of the latest insights regarding VSTLF and STLF. Additionally, an overview of feature selection and customer classification is provided.

In certain applications, including this one under study, the process of load clustering prior to the forecasting step is to be considered. It is essential when tailoring forecasting models to the unique consumption patterns of different consumer groups, by accurately categorizing consumers, forecasters can account for the distinct factors that influence each group. The algorithms regarding load clustering can be divided into five groups: hierarchical, model-based, density-based, grid-based, and partition-based clustering \cite{Cluster1}. Of these, the most popular algorithm is K-means, a type of partition-based algorithm, due to its outstanding computational efficiency on large-scale and high-dimensional load datasets \cite{Cluster2}. A variant of this algorithm has recently been employed to categorise loads, as exemplified by the K-shape classification in \cite{Cluster3}. It is also noteworthy that model-based algorithms have gained prominence, with the emergence of novel approaches such as a Gaussian Mixture Model (GMM) variation in \cite{Cluster4} and a variational autoencoder proposed in \cite{Cluster5}. In some cases, the performance of these models is enhanced by incorporating time-related features, which are extracted by applying domain knowledge. The load patterns are then identified per these features, as illustrated in \cite{Okereke2023, Yu2023}.

A second line of research examines the feature selection process before carrying out the forecasting task, once the different consumers are clustered. The most commonly used technique for feature selection is Principal Component Analysis (PCA), a multivariate statistical analysis technique for data compression and feature extraction that can effectively remove linear correlations between data \cite{PCA1}. It is usually used to reduce dimensionality as in \cite{PCA2}. Another variety of techniques can be employed to calculate the influence of different variables on electric consumption; \cite{Feature1} employs a generalised additive model (GAM) to investigate the impact of various features on residential energy demand, identifying that month, day, and temperature are highly significant in predicting consumption. Another widely used method is the Pearson correlation coefficient, as demonstrated in \cite{feature2, feature3}, whereby the coefficient is employed to ascertain the impact of specific features on the electric load. An additional approach is illustrated in \cite{feature4}, whereby the coarse set of features is initially screened using the Maximal Information Coefficient (MIC), and subsequently, the fine set of key features affecting load forecasting is screened using implementations of Gradient Boosting algorithms, specifically the LightGBM and XGBoost methods, respectively. The corresponding fine set of features and historical loads are input into LightGBM and XGBoost with a robust prediction function for prediction, and the predicted value is employed to rectify the error and complete the load prediction. Other approaches consist of model explainability techniques such as SHAP (SHapley Additive exPlanation). In \cite{Shap_ref}, the technique is employed to reduce feature redundancy and overfitting, after which the selected features are leveraged to forecast day-ahead electricity prices. 

Finally, about the forecasting algorithms, recently there has been a notable shift in focus towards the utilisation of hybrid deep learning algorithms, including CNN-GRU \cite{CNN-GRU}, CNN-LSTM \cite{CNN-LSTM1, CNN-LSTM2}, and even GAN-enhanced models \cite{GAN-enhanced, GAN2}. While these ANNs have become a prominent topic of discussion in the load forecasting literature over the last few years, they present significant challenges when applied to real-world STLF and VSTLF scenarios.  This is largely due to the issue of model overfitting and the exponential increase in complexity associated with high-dimensional data sets \cite{Ceperic}. Additionally, it is important to note that most deep learning models require a substantial number of experiments to identify an optimal configuration \cite{tab_data}.

Consequently, there has been a gradual resurgence of interest in exploring more straightforward regression-based machine learning algorithms. While evaluated, numerous alternative regressors have been demonstrated to exhibit superior efficacy in constructing STLF models compared to ANN. For instance, the Support Vector Regressor (SVR) has been shown to outperform ANN in \cite{ML1}. In \cite{ML2}, the Multiple Linear Regression (MLR), ANN, and SVR are subjected to a comparative analysis, wherein SVR emerges as the most prominent performer. In \cite{ML3}, a sliding window-based LightGBM model has been shown to outperform the popular LSTM deep learning algorithm. 

An additional illustration of the efficacy of tree-based gradient boosting algorithms can be observed in \cite{XGB}, where the model is a combined method based on similar days matching and the XGBoost method for short-term load forecasting on holidays. This approach is proposed as a means of addressing the problem of load forecast errors on such days.  One challenge associated with these gradient-boosting algorithms is the accurate specification of hyperparameters. This is addressed in \cite{bayesian2} through the use of Bayesian Optimization, which is employed to optimise the hyperparameters for a LightGBM model prior to performing the forecasting.

\section{Methodology}
\begin{figure}[!ht]
    \centering 
    \includegraphics[width=0.9\textwidth]{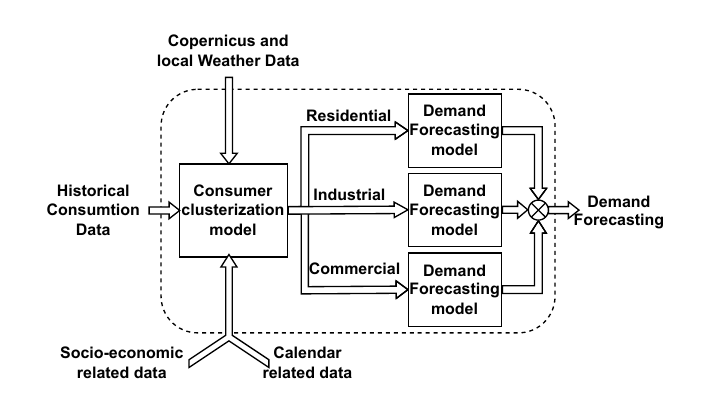}
    \caption{Framework of the load demand forecasting system}
    \label{fig: Framework}
\end{figure}

In this section, we describe the developed load forecasting method in detail. The overarching framework of the load demand forecasting system is depicted in Figure \ref{fig: Framework}.
This system begins by collecting a variety of input data sources, further discussed in Section 4, including historical energy consumption records provided by the stakeholders in the project, weather-related data obtained from the ERA5 dataset and regional meteorological stations, as well as socio-economic indicators and calendar-related data.

The first step in the forecasting process involves classifying customers into distinct categories—residential, industrial, and commercial—based on their energy consumption features and the characteristics of their respective data. This classification is essential as each customer category exhibits unique behavior and demand drivers that must be considered when forecasting load.

Once customers are classified, specific demand forecasting models are tailored and applied to each customer category. The model selection is critical, as it ensures that the most appropriate forecasting techniques are employed, considering each category's unique attributes and influencing factors. As a final step, the consumption forecasts for different customers at the same location are aggregated. By utilizing a targeted approach for each customer type, the system enhances the accuracy and reliability of the load forecasts, ultimately supporting better grid management and planning.

\subsection{Customer classification}
One possible approach would be to implement a pattern recognition algorithm using time-domain features, such as the K-Means clustering algorithm, as demonstrated in \cite{Okereke2023, Yu2023}, to distinguish between different consumption patterns. Furthermore, if we had access to additional consumer characteristics such as household appliances data, consumers behaviour, or the size of the companies in the case of industrial and commercial customers, we could apply more sophisticated methods, as described in \cite{Wang2021} or \cite{Dudek2018}. However, all the collected data is completely anonymous and only the customers' type and locations are available as input data, thus, having the ground truth, we developed a classifier based on descriptive characteristics and domain knowledge.

\begin{enumerate}
    \item If the consumption during the holidays \((C_{H})\) is less than half the consumption during the working days \((C_{W})\), and the mean consumption on Saturdays \((\hat{C}_{6})\) is less than twice the mean consumption on Sundays \((\hat{C}_{7})\), the algorithm classifies the dataset as \textit{industrial} consumers:

\begin{equation}
    \begin{aligned}
    C_{H} < \frac{C_{W}}{2}
    \\
    \hat{C}_{6} < 2 \hat{C}_{7} 
    \end{aligned}\label{Rule 1}
\end{equation}
    The first condition differentiates industrial and commercial consumers from residential ones, as they completely stop or reduce consumption to a minimum during holidays. The second condition separates the industrial consumers from the commercial ones, as the industrial consumers normally reduce their consumption on Saturdays too, in contrast to the commercial consumers, who tend to work on Saturdays.\\
    
    \item Once the industrial consumers are separated, a differentiation between the commercial and the residential consumers is sought. The dataset is classified as \textit{commercial} if the standard deviation of mean consumption at different hours exceeds 0.1, and if mean consumption on holidays is lower than on workdays. The first condition identifies commercial datasets, which have consistent consumption peaks at specific hours, leading to a higher standard deviation compared to the more irregular residential consumption patterns. The second condition prevents misclassification of residential data that fits the first condition. Additionally, if mean consumption on Saturdays is more than 1.5 times that on Sundays, the dataset is also classified as \textit{commercial}.

\begin{equation}
    \begin{aligned}
    std(\hat{C}_i) > 0.1
    \\
    \hat{C}_{H} > \hat{C}_{W} 
    \\
    \hat{C}_{6} > 1.5 \hat{C}_{7} 
    \end{aligned}\label{Rule 2}
\end{equation}
    This condition is similar to the first applied rule, which compares the consumption during workdays and holidays, as well as the consumption during Saturdays and Sundays, which helps to not classify commercial data as residential.\\

    \item If the dataset does not fall into the two previous categories, i.e., does not meet the conditions established by equations (1) and (2), it is classified as \textit{residential} consumer.
\end{enumerate}

\subsection{Model training}
\begin{figure}[!ht]
    \includegraphics[width=0.9\textwidth]{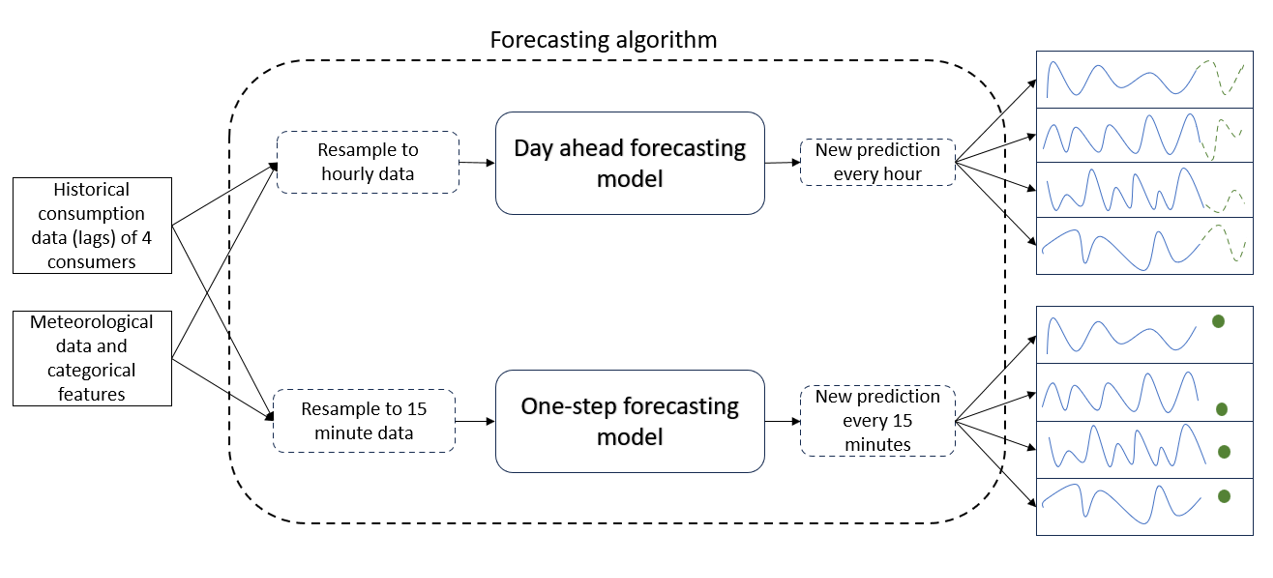}
    \caption{Design of the forecasting algorithms}
    \label{fig:Design}
\end{figure}

This section aims to state the process of designing the forecasting models for each consumer type, once they are classified; the reasoning behind model selection, parameter selection, and training process are also discussed. In Figure \ref{fig:Design} the diagram of the proposed algorithm architecture is shown, where it can be seen that based on historical consumption data and the most relevant features, further detailed in Section 3.2.1, and after being resampled adequately, each forecasting algorithm needs to perform two distinct forecasting tasks, one is a one-step forecast 15 minutes ahead in the future, and the other performs a prediction of the next 24h at each time step of the testing data, i.e., hourly.

A total of six forecasting models have been designed, with two models for each type of consumer (day ahead and 15 minutes ahead). It is also worth noting that before training the models, a baseline should be established for each model, to compare the results and metrics obtained by the forecasting algorithms with the baseline's metrics. In the case of the one-step forecast, the baseline is determined by repeating the last lag, i.e., the power demand from 15 minutes earlier, and comparing it with the current power demand. As for the day ahead forecast, the baseline is established by repeating the previous 24 samples (the power demand from the previous day) and comparing it to the current day’s power demand.

\subsubsection{Feature selection}

Regarding the input features of the forecasting models, a feature importance analysis is conducted for each consumer type to guarantee that only the most relevant features are incorporated as inputs. These features include:
\begin{itemize}
    \item Calendar-related categorical features such as the month, day of the week, and time.
    \item Holiday binary feature.
    \item Weather data, in particular hourly temperature and humidity data. 
    \item Socio-economical features, specifically the total population, population density, the territorial socio-economic index, and the gross disposable household income.
\end{itemize}

It is imperative to note that all of these features are included as future covariates. This implies that they must be known in advance, with the requirement being to be known 15 minutes into the future and 24 hours ahead. Furthermore, it is essential that the features are correctly handled to ensure that the models receive the data in the most appropriate way, for which a preprocessing step was included. The weather data was standardized between 1 and -1, while the encoding of calendar-related features and the holiday feature is model-dependent, varying according to the forecasting model in use. For tree-based models (LightGBM, XGBoost and RF), these variables were categorically encoded; for the rest (MLP, SVR and LSTM), they were one-hot encoded.

The feature selection process, as depicted in Algorithm \ref{alg:1}, is a Backward Elimination process that employs the training of LightGBM and XGBoost models with and without each feature. LightGBM and XGBoost were selected for this study due to their strong performance with limited data, as well as their ease of training, reliability, and speed, \cite{LGBM1,bayesian2}. These methods offer significant advantages, including robust regularization capabilities and fine-tuned control over model complexity are particularly valuable when working with smaller datasets or when overfitting is a significant concern.

However, we will also incorporate SHAP to demonstrate the model-agnostic nature of our feature selection process. SHAP provides consistent, interpretable explanations of feature importance by assigning Shapley values that represent each feature's contribution to the model's predictions. Since SHAP is model-agnostic, it enables us to evaluate feature importance independently of any specific model structure. By ranking features based on their SHAP values, we ensure that our feature selection is robust across different models, supporting a transparent and objective approach that avoids reliance on any particular algorithm's behavior. This enhances the reliability and interpretability of the selected features.

\begin{algorithm}
\caption{Feature selection process}\label{alg:1}
\begin{algorithmic}[1]
    \State $S = \{s_1,s_2,...,s_n\}$  \Comment{All features}
    \State $S' = \{s'_1,s'_2,...,s'_m\}$ \Comment{Important Features}
        \State Train single feature LightGBM and XGBoost models
        \State Calculate error of the model, $e_{u}$
        \For{$s_{i}$ in $S$}
            \State Train LightGBM and XGBoost with $s_i$ as a feature 
            \State Calculate the error of the model $e_i$
            \If{$e_i \gtrsim   e_u$}
                \State $s_i \in S'$
            \Else 
                \State $s_i \notin S'$
            \EndIf
        \EndFor
\end{algorithmic}
\end{algorithm}

\subsubsection{Model selection}
Once the most relevant features have been identified, a series of machine learning models will be evaluated, extending beyond the previously employed LightGBM and XGBoost. The following machine learning models will be tested: MLP (Multilayer Perceptron), LSTM (Long Short-Term Memory), RF (Random Forest), and SVR (Support Vector Regression). These algorithms were selected as they have consistently demonstrated their efficiency in STLF under different circumstances \cite{ML1, ML2, ML3}. The majority of studies present either a comparison between different deep learning models (LSTM, MLP, etc.) \cite{CNN-LSTM2} or a comparison between machine learning models (LightGBM, SVR, RF, etc.) \cite{ML1}.

These algorithms represent diverse approaches to time-series forecasting including gradient boosting (LightGBM and XGBoost), feedforward neural network (MLP), recurrent neural networks (LSTM) and a supervised max-margin model (SVR). The purpose of this comparison is to ascertain which technique performs better for which consumer type.

\subsubsection{Bayesian hyperparameter optimization}
In order to optimize the performance of the model in the validation set, it is often necessary to exhaustively explore a vast number of different combinations of hyperparameter values, which is a time-consuming process.

Bayesian optimization is a suitable approach for addressing the aforementioned issue, which can be divided into two main components: 1) a probabilistic surrogate model, which is employed to approximate the current black box objective function; and 2) an acquisition function, which is used to identify the optimal solution under the current data conditions \cite{bayesian}. By reasoning with past search results and attempting more promising values, Bayesian optimization reduces the time required for hyperparameter search and facilitates the identification of optimal values. The Bayesian hyperparameter optimization is represented in Equation \ref{Bayesian}.
\begin{equation}
    \begin{aligned}  
    x^{*} = argmin f(x) \implies   x \in    \chi 
    \end{aligned}\label{Bayesian}
\end{equation}
where, $f(x)$ denotes the target value to be minimized for evaluation on the validation set, which is the root mean squared error (RMSE) in this case. $\chi$ is the domain of hyperparameters \cite{bayesian2}.

In our study a Tree-structured Parzen Estimator (TPE) \cite{TPE} is used as a subrogate model which is used to estimate the densities of good hyperparameters and bad hyperparameters.

\subsection{Evaluation criteria}\label{subsec. 3.3}
This subsection describes the evaluation criteria that were derived from input provided by stakeholders involved in the Horizon Europe project, composed of Energy Services Companies (ESCOs), DSOs, and big industrial consumers.

\subsubsection{Classification algorithm evaluation}
The Key Performance Indicator (KPI) set for the classification algorithm is set in terms of accuracy.
Drawing on results from the literature \cite{Dudek2018, Wang2021}, we have set a threshold requiring the algorithm to achieve at least 75\% accuracy in classifying the different customer types.

\subsubsection{Forecasting model evaluation}
The KPIs established for the forecasting models are also expressed in terms of accuracy; the 15-minute-ahead forecast needs to achieve an accuracy of at least 80\%, and at least 85\% in the case of the day-ahead forecast. Error metrics that provide information similar to accuracy in a prediction include MAE (Mean Absolute Error) and MAPE (Mean Absolute Percentage Error), shown in (\ref{Metrics}):

\begin{equation}
    \begin{aligned}
    MAE = \frac{1}{n} \sum_{i=1}^{n} \left| s_{i} - s_{i}^* \right|\\
    MAPE = \frac{100\%}{n} \sum_{i=1}^{n} \left|\frac{s_{i} - s_{i}^*}{s_{i}} \right|
    \end{aligned}\label{Metrics}
\end{equation}
where $s_i$ and $s_i^*$ denote the actual and forecasted load values respectively.

Accuracy measures how closely forecasted values match actual values, typically expressed as a percentage. MAPE, which is scale-independent and interpretable as a percentage, is useful for comparing forecasts across datasets, while MAE provides error in the same units as the data and is robust to outliers. Optimizing MAPE can lead to demand underestimation, whereas optimizing MAE aims to balance errors, which can be problematic with extreme load variations such as in industrial consumers. Low load values can challenge MAE and MAPE evaluation, as occurs in residential consumers, \cite{Metrics}. To address instances of significant error despite low average metrics, we use a quantitative score to measure the frequency of MAPE and MAE falling below a set threshold, highlighting periods of model underperformance potentially due to unaccounted seasonality or complex factors.

It is also important to note that in order to accurately assess the performance of the day-ahead forecaster, a production scenario is generated by predicting the next 24h at each time step of the testing data, rather than a single prediction every 24h. This approach allows for the creation of overlapping predictions, which can then be individually evaluated to provide a more comprehensive assessment of the forecaster's accuracy.

\section{Case study}
In this section, we will introduce the specific case study under investigation and provide a detailed description of the datasets utilized in the analysis, including their sources, and the types of data collected. This section aims to establish a clear understanding of the foundational elements of the study, setting the stage for subsequent discussions of the results.

\subsection{Historical load data}
\begin{figure}[h]
    \centering 
    \includegraphics[width=0.65\textwidth]{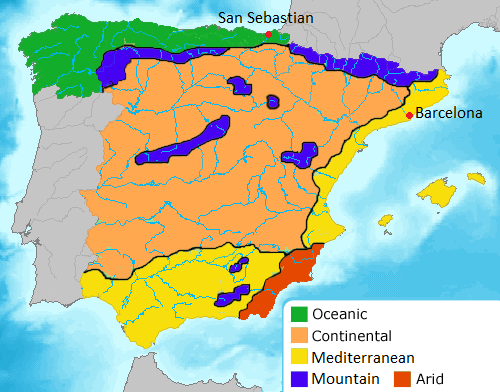}
    \caption{Different climatic zones in Spain}
    \label{fig: clima}
\end{figure}

The available data set comprises 15-minute interval data for two years of power consumption for real use cases located in Spain, of four \textit{industrial} and two \textit{commercial} consumers, data provided by an SME specialized in global electrical solutions, control, electrical maintenance, and data analysis. Moreover, a DSO also provided data for five \textit{residential} consumers. Spain has different climatic zones, as depicted in Figure \ref{fig: clima}, and as we are going to take weather-related variables into account, it is important to know the location of each customer. In our case, specifically, the \textit{industrial} and \textit{commercial} consumers are situated in the Basque Country, near San Sebastian, on the coast of the Bay of Biscay, dominated by a warm, humid, and wet oceanic climate. On the other hand, the \textit{residential} consumers are located in Catalonia, near Barcelona, in the northeast of Spain, with a Mediterranean climate, Figure \ref{fig: clima}. However the differences in the climate, these two regions have similar GDPs and socio-economic characteristics.

\subsection{Weather data}
Weather-related data is utilized as input features for the forecasting models, with the impact of this data on forecasting accuracy analyzed for each consumer type. Different consumers have varying relationships with specific weather variables that affect their energy usage and environmental performance. According to \cite{Amin2024} sixteen parameters are widely used for different energy applications; with dry-bulb temperature [°C] being the most prevalent, followed by humidity [\%], both of which are essential and highly influential across most energy applications throughout their life cycle. Wind speed [$m/s$] and global solar radiation [$W/m^2$] are also frequently used, particularly in studies focused on specific buildings \cite{Luo2021, Salata2020}.

However, in our study, since we lack information on specific buildings, we will focus solely on humidity and dry-bulb temperature. Although these variables were initially obtained from the ERA5  dataset \cite{era5}, the resulting forecasting accuracy was not sufficiently high due to the spatial resolution limitations of the ERA5 and CAMS free services. To enhance the accuracy of the forecasts, we have incorporated additional data from the national meteorological agency, specifically Aemet \cite{Aemet}.

\subsection{Socio-economic data}
Socio-economic factors play a significant role in demand forecasting across different regions. However, for industrial and commercial consumers, where the businesses are of similar size and located in the same region, these factors are excluded from the analysis as they provide limited additional information. Thus, this data is only used for residential consumers, which correspond to different towns near Barcelona, and for each of the towns, the following data was collected the regional institute of statistics, Idescat \cite{Idescat}: I) total population, II) population density, III) TSI (territorial socio-economic index), and IV) GDHI (Gross Disposable Household Income). Income levels impact the affordability and use of energy-intensive appliances, with higher-income households more likely to invest in energy-efficient technologies, reducing overall consumption \cite{delReal, Hu2020}. Additionally, household size, employment rates, and daily routines influence energy consumption, with demand fluctuating based on periods of activity and rest.

\section{Results}
This section presents a comprehensive discussion of the results obtained from the customer classification, feature selection, and forecasting models. We will analyze the effectiveness of the classification process in accurately grouping customers into their respective categories, followed by an evaluation of the feature selection process employed to identify the most relevant variables influencing demand. Finally, we will conduct a detailed examination of the performance of the forecasting models, each of which has been specifically tailored to the respective customer groups. In cases where the results do not meet the established standards, alternative approaches will be introduced and compared, demonstrating their contribution to enhancing previous predictions. The insights derived from these analyses will help to inform the broader implications of the study.

\subsection{Customer Classification Results}
\begin{figure}[h]
    \centering 
    \includegraphics[width=0.65\textwidth]{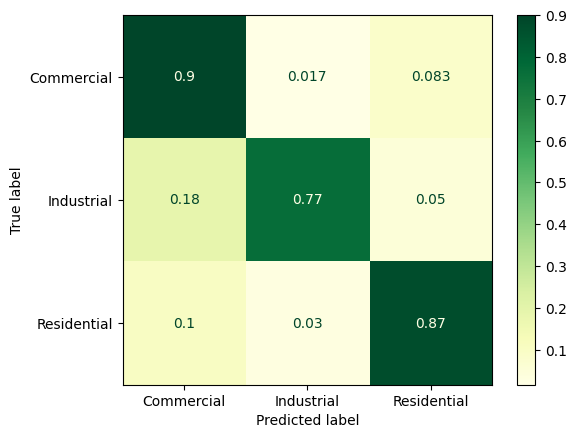}
    \caption{Confusion matrix of dataset classificator}
    \label{fig: classif}
\end{figure}

Results from the rule-based classification algorithm are shown in the confusion matrix of Figure \ref{fig: classif}. There is a significant imbalance in the number of available datasets for each category, which poses a challenge for conducting comprehensive analyses. To address this issue, we have divided the underrepresented datasets into more subsets (6 subsets in the case of residential datasets, and 30 in industrial and commercial datasets) to ensure a balanced representation across different categories. This approach allows for a more equitable distribution of data, enhancing the robustness of the analysis while reducing potential biases caused by the overrepresentation of certain dataset types. By doing so, we aim to provide a more thorough and accurate evaluation that accounts for the inherent data discrepancies. The algorithm performs correctly in overall 85\% of the cases. The industrial datasets are classified correctly 77\% of the time, accuracy in residential datasets is 87\%, and accuracy in commercials is 89\%. These values comply with the established KPIs, which state that the classification algorithm must cluster consumers (residential, commercial, industrial) with a precision of> 75\%.

\subsection{Feature Selection Results}
\subsubsection{Industrial consumers}
The results presented in Table \ref{tab1}, about the day-ahead forecasting models, indicate whether the model's performance improves (highlighted in green) or worsens (highlighted in red) in terms of MAPE and MAE. These comparisons are made concerning the XGBoost and LightGBM methods following the introduction of various features

These results indicate that weather does not significantly impact the power consumption of industrial consumers. Additionally, neither does the 15-minute-ahead forecaster. These results are coherent with the fact that the main power consumption source for industrial consumers is their heavy machinery, and although their overall consumption may be affected by temperature and humidity, this influence is negligible compared to the demand from the machinery.

On the other hand, the inclusion of \textit{calendar information }(hour, day of the week, and month) is beneficial for the forecasting models as power consumption exhibits seasonal patterns. This allows the model to capture these seasonalities more accurately. This is again coherent, considering that the factories’ production is directly related to the day of the week, months, and an hours-related schedule.

Finally, regarding the binary \textit{holiday variable}, the forecasters show a significant improvement in accuracy, which is an expected result, as the majority of factories close up during holidays, or at least, work at a minimum capacity, so their consumption is directly altered by that factor.

\begin{figure}[h]
    \centering 
    \includegraphics[width=0.8\textwidth]{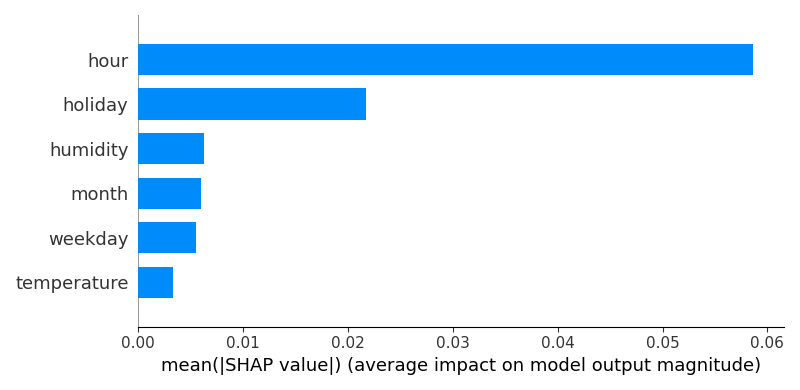}
        \caption{Shap values for the industrial model features}
    \label{fig: shap_ind}
\end{figure}

We will compare these results with SHAP analyses to ensure that the feature importance and performance improvements are model-agnostic, aiming to confirm that the observed trends hold consistently across different models, providing greater confidence in the robustness and generalizability of our findings. With Figure \ref{fig: shap_ind}, we can confirm that the most important features are the hour and holiday data, whereas the weather data and month/weekday features do not impact as much.

\begin{table}[!h]
\centering
\caption{Error change of industrial models to perform feature selection \label{tab1}}
{\begin{tabular}{|c|c|c|cccc|}
\toprule
Feature &Model &Metric &Ind.1 &Ind.2 &Ind.3 &Ind.4\\
\midrule
               &XGB  &MAPE &\cellcolor{green!25}-1.1 &\cellcolor{red!25}0.5 &\cellcolor{red!25}0.3 & 0.0 \\
Temp. \&    &     &MAE  &\cellcolor{green!25}-0.8 &\cellcolor{red!25}1.5 &\cellcolor{red!25}0.7 &\cellcolor{green!25}-0.1 \\
Humidity   &LGBM &MAPE &\cellcolor{green!25}-0.7 &\cellcolor{green!25}-0.1 &\cellcolor{green!25}-1.3 &\cellcolor{green!25}-0.1 \\
               &     &MAE  &\cellcolor{green!25}-0.8 &\cellcolor{green!25}-0.1 &\cellcolor{green!25}-0.7 &\cellcolor{green!25}-0.2 \\
\midrule
               &XGB  &MAPE &\cellcolor{green!25}-30.9 &\cellcolor{green!25}-8.1 &\cellcolor{green!25}-1.0 &\cellcolor{green!25}-1.2 \\
Calendar       &     &MAE &\cellcolor{green!25}-6.0 &\cellcolor{green!25}-5.4 &\cellcolor{green!25}-2.7 &\cellcolor{green!25}-2.0 \\
Features       &LGBM &MAPE &\cellcolor{green!25}-28.7 &\cellcolor{green!25}-6.4 &\cellcolor{green!25}-5.5 &\cellcolor{green!25}-1.5 \\
               &     &MAE &\cellcolor{green!25}-5.9 &\cellcolor{green!25}-6.3 &\cellcolor{green!25}-0.4 &\cellcolor{green!25}-1.6 \\
\midrule
               &XGB  &MAPE &\cellcolor{green!25}-14.1 &\cellcolor{green!25}-11.0 &\cellcolor{green!25}-3.1 &\cellcolor{green!25}-1.1 \\
Holiday        &     &MAE &\cellcolor{green!25}-3.8 &\cellcolor{green!25}-5.7 &\cellcolor{green!25}-1.4 &\cellcolor{green!25}-1.7 \\
Indicator      &LGBM &MAPE &\cellcolor{green!25}-20.4 &\cellcolor{green!25}-10.8 &\cellcolor{green!25}-4.4 &\cellcolor{green!25}-0.8 \\
               &     &MAE &\cellcolor{green!25}-4.9 &\cellcolor{green!25}-5.5 &\cellcolor{green!25}-2.1 &\cellcolor{green!25}-1.2 \\
\bottomrule
\end{tabular}}{}
\end{table}

\subsubsection{Commercial consumers}
In the same way, the results gathered in Table \ref{tab3} indicate that although slightly, all the features improve the accuracy of the models, however, this improvement is not always significant. The case of \textit{calendar information} and the \textit{holiday variable} is similar to the industrial consumers' case, as their consumption has a seasonality related to the calendar schedule and the holidays.

Nevertheless, in this case, unlike in the industrial case, the weather-related variables do affect the commercially related behavioral patterns of the people. On the one side, reaching the temperature comfort levels inside a commercial building will directly affect power consumption. On the other hand, it also has to do with the encouragement of the people to go to these commercial sites, which in turn again directly impacts the temperature comfort levels.

\begin{figure}[h]
    \centering 
    \includegraphics[width=0.8\textwidth]{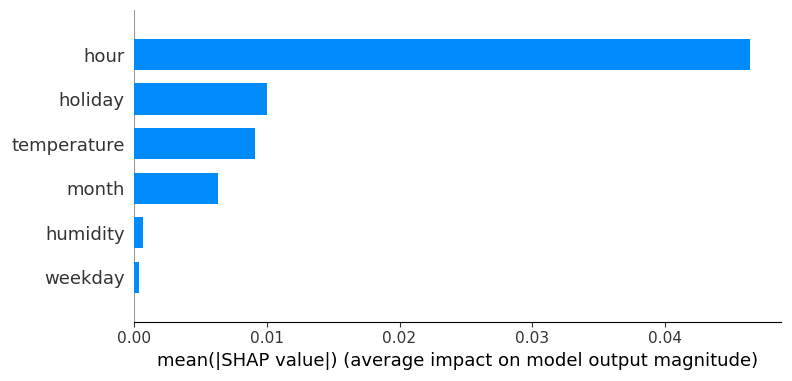}
        \caption{Shap values for the commercial model features}
    \label{fig: shap_com}
\end{figure}

Again, we will compare these results with SHAP analyses. With Figure \ref{fig: shap_com}, we confirm that the most important features are the hour and holiday data. We can furthermore elaborate on the results in Table \ref{tab3} related to weather variables, as temperature is the important one, while humidity is virtually irrelevant.

\begin{table}[!h]
\centering
\caption{Error change of commercial models to perform feature selection \label{tab3}}
{\begin{tabular}{|c|c|c|cc|}
\toprule
Feature &Model &Metric &Com.1 &Com.2\\

\midrule
               &XGB  &MAPE &\cellcolor{green!25}-0.12 &\cellcolor{green!25}-2.21 \\
Temp. \&   &     &MAE  &\cellcolor{green!25}-0.01 &\cellcolor{green!25}-0.14 \\
Humidity   &LGBM &MAPE &\cellcolor{green!25}-0.9  &\cellcolor{green!25}-1.43 \\
               &     &MAE  &\cellcolor{green!25}-0.54 &\cellcolor{green!25}-0.12 \\
\midrule
               &XGB  &MAPE &\cellcolor{green!25}-1.59 &\cellcolor{green!25}-6.36 \\
Calendar       &     &MAE  &\cellcolor{green!25}-0.46 &\cellcolor{green!25}-0.22 \\
Features       &LGBM &MAPE &\cellcolor{green!25}-1.35 &\cellcolor{green!25}-4.23 \\
               &     &MAE  &\cellcolor{green!25}-0.78 &\cellcolor{green!25}-0.10 \\
\midrule
               &XGB  &MAPE &\cellcolor{green!25}-1.08 &\cellcolor{green!25}-0.92 \\
Holiday        &     &MAE  &\cellcolor{green!25}-1.01 &\cellcolor{green!25}-0.15 \\
Indicator      &LGBM &MAPE &\cellcolor{green!25}-2.44 &\cellcolor{green!25}-2.36 \\
               &     &MAE  &\cellcolor{green!25}-1.35 &\cellcolor{green!25}-0.10 \\
\bottomrule
\end{tabular}}
\end{table}

\subsubsection{Residential consumers}
Finally, the results for the residential consumers gathered in Table \ref{tab2} show that the impact of the different features is not quite significant, although intuitively, we would expect a higher impact of the \textit{temperature} and \textit{humidity}. We would expect residential consumers to be more likely to adjust their energy usage based on temperature changes, leading to spikes in demand for heating or cooling, but in this case study that relationship does not seem so direct.

Regarding the inclusion of \textit{calendar information }(hour, day of the week, and month) and the binary \textit{holiday variable}, they too have a very small impact on the predictions. 

\begin{figure}[h]
    \centering 
    \includegraphics[width=0.8\textwidth]{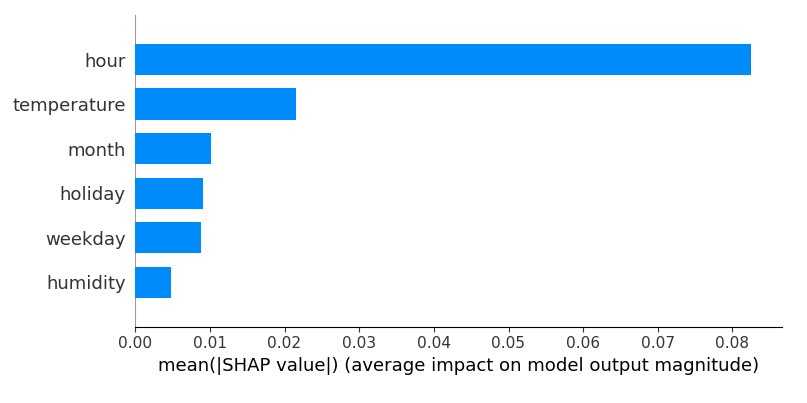}
        \caption{Shap values for the residential model features}
    \label{fig: shap_res}
\end{figure}

However, if we compare the results with SHAP analyses in Figure \ref{fig: shap_res}, we see that the most important feature is again the hour. As expected, temperature ranks as the second most important feature, while other calendar-related data, including holiday information—which is the second most important feature for other consumers—appear to be less relevant in this case.

\begin{table}[!h]
\centering
\caption{Error change of residential models to perform feature selection \label{tab2}}
{\begin{tabular}{|c|c|c|ccccc|}
\toprule
Feature &Model &Metric &Res.1 &Res.2 &Res.3 &Res.4 &Res.5\\

\midrule
               &XGB  &MAPE &\cellcolor{green!25}-0.9 &\cellcolor{green!25}-0.55 &\cellcolor{green!25}-0.9 &\cellcolor{green!25}-1.0 &\cellcolor{green!25}-0.73 \\
Temp.\&   &     &MAE  & 0.0 & 0.0 & 0.0 &\cellcolor{green!25}-0.01 & 0.0 \\
Humidity   &LGBM &MAPE &\cellcolor{green!25}-0.48 &\cellcolor{green!25}-0.47 &\cellcolor{green!25}-0.01 &\cellcolor{green!25}0.04 &\cellcolor{green!25}-0.1  \\
               &     &MAE  &\cellcolor{green!25}-0.01& 0.0 & 0.0 & 0.0 & 0.0 \\
\midrule
               &XGB  &MAPE &\cellcolor{green!25}-1.24 &\cellcolor{green!25}-1.13 &\cellcolor{green!25}-1.2 &\cellcolor{green!25}-2.55 &\cellcolor{green!25}-3.38 \\
Calendar       &     &MAE &\cellcolor{green!25}-0.01 &\cellcolor{green!25}-0.01 &\cellcolor{green!25}-0.02 & 0.0 & 0.0 \\
Features       &LGBM &MAPE &\cellcolor{green!25}-1.31 &\cellcolor{green!25}-1.97 &\cellcolor{red!25}0.1 &\cellcolor{green!25}-1.4 &\cellcolor{green!25}-1.87 \\
               &     &MAE & 0.0 &\cellcolor{green!25}-0.02&\cellcolor{green!25}-0.01&\cellcolor{green!25}-0.01&\cellcolor{green!25}-0.01 \\
\midrule
               &XGB  &MAPE &\cellcolor{red!25}0.56 & 0.0 &\cellcolor{green!25}-0.59 &\cellcolor{green!25}-0.84 &\cellcolor{green!25}-0.44 \\
Holiday        &     &MAE &\cellcolor{red!25}0.01 & 0.0 & 0.0 & 0.0 & 0.0 \\
Indicator      &LGBM &MAPE &\cellcolor{green!25}-0.49 &\cellcolor{green!25}-0.82 & 0.0 &\cellcolor{green!25}-0.1 & 0.0 \\
           &     &MAE & 0.0 & 0.0 & 0.0 & 0.0 & 0.0 \\
\bottomrule
\end{tabular}}{}
\end{table}

\subsection{Model Selection Results}

In this section, the models' results are going to be evaluated and discussed, based on the KPIs established by the stakeholders of the Horizon project, which are the accuracies to be achieved. Although it can be challenging to evaluate trained models regarding their accuracy precisely, it can be done through diverse metrics that represent different error characteristics, such as the MAPE and MAE, which are valuable tools, especially when used in conjunction with quantitative assessments. The overall MAPE and MAE results gathered in Tables \ref{tab_results_24} and \ref{tab_results_15}, in \ref{sec_append}, are therefore combined with a quantitative score, previously introduced in Section \ref{subsec. 3.3}.

To calculate said quantitative score and evaluate the performance of the model, thresholds need to be established for MAPE and MAE. MAPE provides an overall assessment of the model's predictive accuracy; being the target accuracy of over 80\% for day-ahead forecasts and over 85\% for 15-minute-ahead forecasts, we will set the MAPE threshold at 20\% for day-ahead forecasts and 15\% for 15-minute-ahead forecasts. These thresholds will apply to all consumer types, as MAPE is scale-independent.

On the other hand, while MAE also measures the average magnitude of errors between predicted and actual values across an entire dataset, it is scale-dependent, meaning it will vary for each consumer. To establish the threshold for MAE, we will allow a 20\% error for day-ahead forecasts and a 15\% error for 15-minute-ahead forecasts. For each consumer, the mean value is obtained and the corresponding 20\% and 15\% values are calculated to establish the MAE thresholds, gathered in Tables \ref{paInd_MAE}, \ref{paComm_MAE} and \ref{paRes_MAE}.

Finally, we will set the target quantitative score values to align with the accuracy benchmarks to be achieved. This approach ensures that the model is sufficiently robust and effective in consistently meeting the desired MAPE and MAE thresholds over extended periods. These values also will be valid for all types of consumers.

\subsubsection{Industrial consumers}

\begin{figure}[h]
    \centering 
    \includegraphics[width=0.8\textwidth]{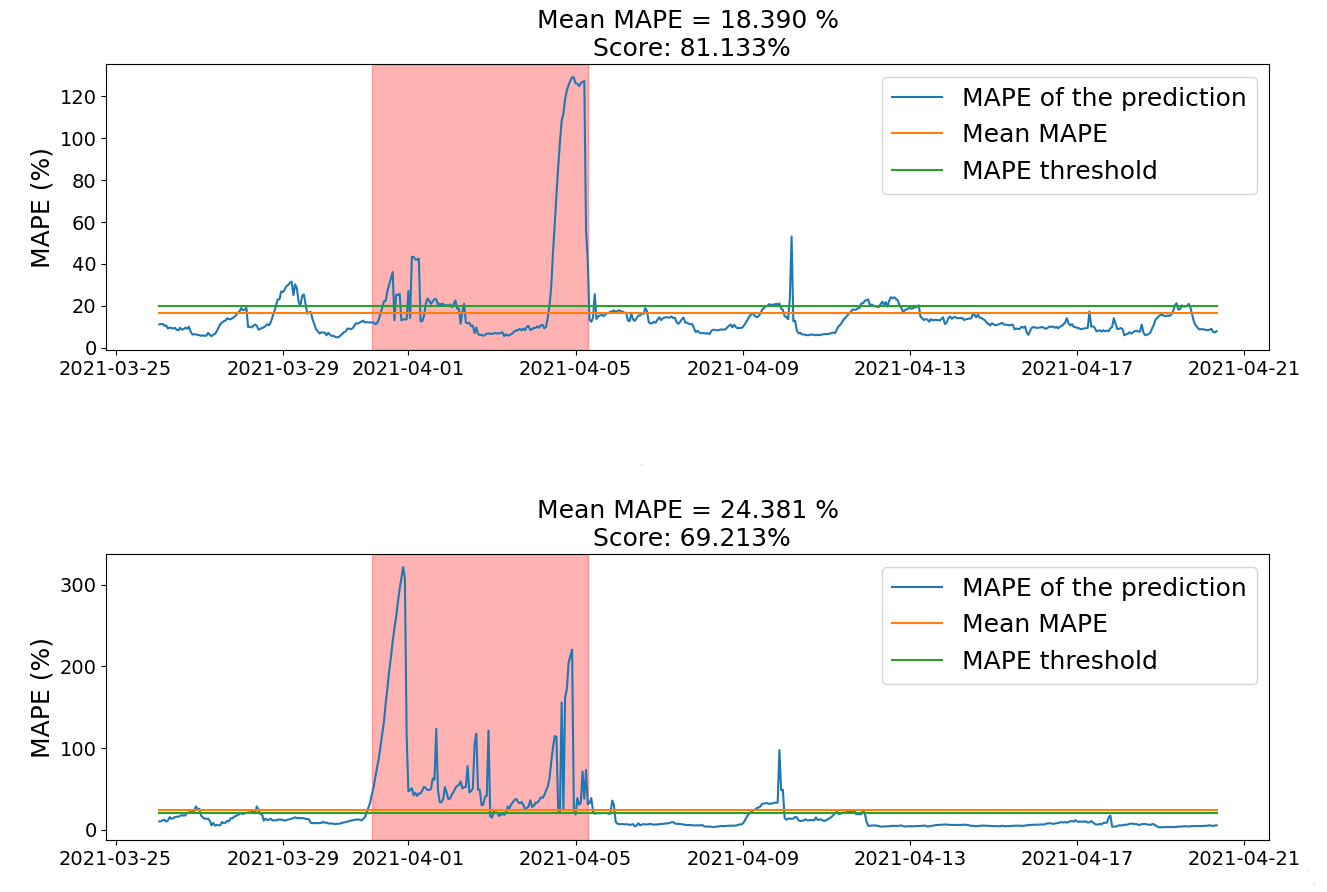}
    \caption{Evolution of MAPE in different industrial consumers}
    \label{fig: MAPE-Ind}
\end{figure}

For industrial consumers, the chosen model is LightGBM, as it is the best-performing one according to the results in section (\ref{sec_append}). However, if we plot the MAPE of the day-ahead forecasters over one month of testing data, as in the examples in Figure \ref{fig: MAPE-Ind}, it can be seen that it underperforms when the dates correspond to holidays: the marked area relates to the Easter holidays of 2021.

This issue primarily arises due to, despite incorporating holiday data as an exogenous variable, the forecaster occasionally misclassifies a holiday as a working day, resulting in incorrect predictions of peak power demand. Additionally, industrial consumers exhibit distinctly bimodal behavior, with significant differences in consumption between weekdays and weekends or holidays. These discrepancies mean that if weekday consumption is predicted incorrectly, the resulting error is much larger compared to other consumer types, such as residential consumers, where consumption values do not vary as drastically.

\begin{figure}[h]
    \centering    \includegraphics[width=0.65\textwidth]{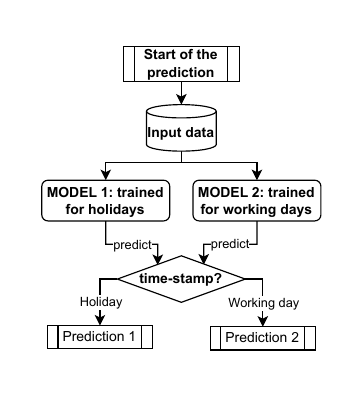}
    \caption{Proposed approach}
    \label{fig: fusion}
\end{figure}

\paragraph{\textbf{Proposed approach}}
To cope with this problem, another forecasting approach is proposed below, depicted in Figure \ref{fig: fusion}, combining, or fusing, two distinct models, both LightGBM-based, to perform the prediction: we train two distinct forecasting models, one only for holidays and the other one for working days. Then, at prediction time, both models perform the prediction but as the holiday information of the next 24 hours is known, only the prediction for the correct model is taken.

\begin{figure}[h]
    \centering 
    \includegraphics[width=0.8\textwidth]{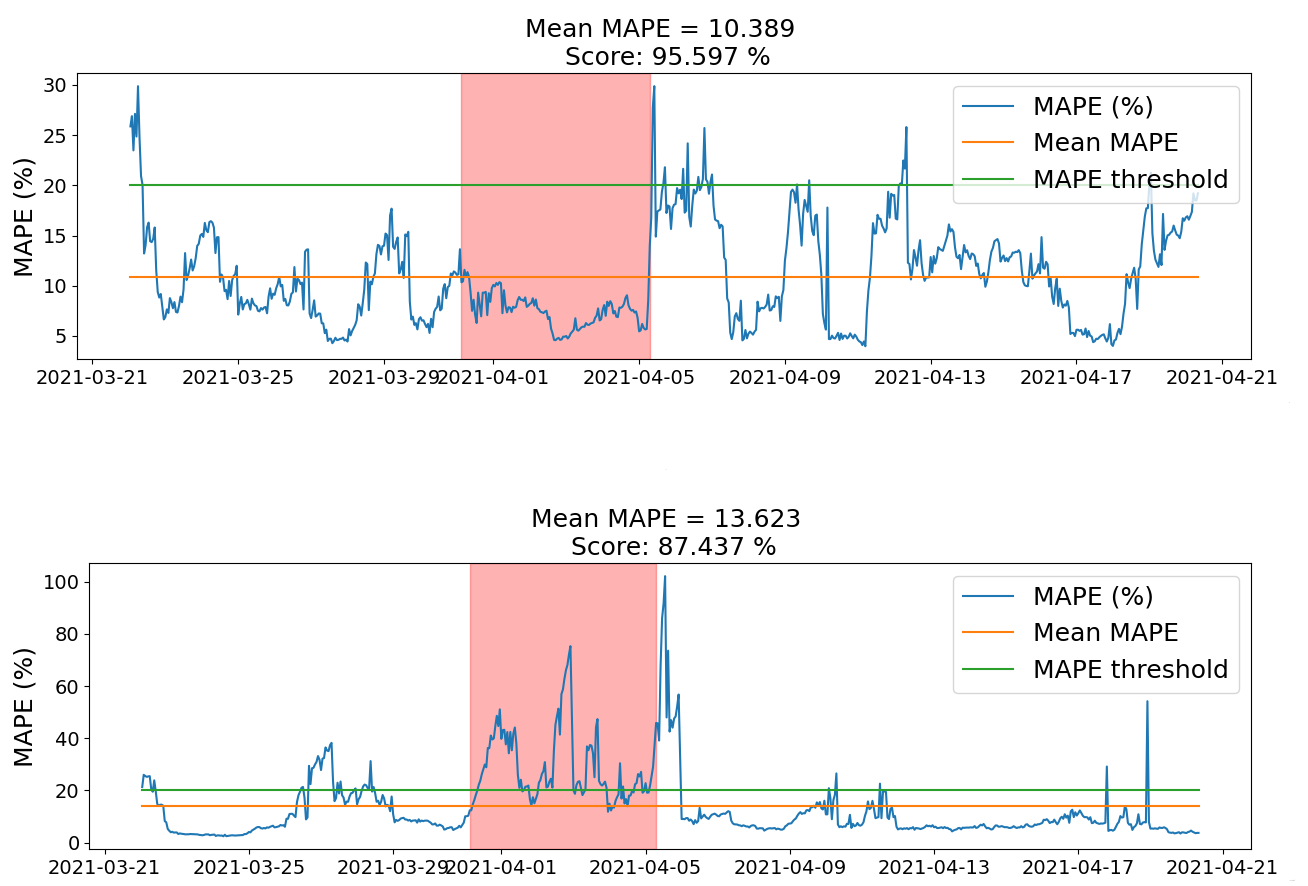}
    \caption{Evolution of MAPE in different industrial consumers after applying proposed approach}
    \label{fig: MAPE-Ind2}
\end{figure}

In Figure \ref{fig: MAPE-Ind2}, it can be appreciated that, although some peaks can still be appreciated, a significant improvement is achieved in terms of MAPE, and how it holds over the tested month, even in holidays.

In Tables \ref{paInd_MAPE} and \ref{paInd_MAE} results from applying a single model versus applying the proposed fusion approach are compared, both in terms of MAPE, MAE, and the quantitative score reached.

The results indicate that, although the MAPE and MAE values are below the established thresholds even when applying a single model, the quantitative score obtained does not meet the required standards, suggesting that the model lacks robustness. In contrast, the model fusion approach produces better outcomes overall, significantly lowering both MAPE and MAE while also increasing the quantitative score values. This improvement reflects the enhanced overall performance of the model over extended periods, demonstrating its superior effectiveness and stability.

\begin{table*}[!ht]
\centering
\caption{Comparison of MAPE results for the industrial consumers with the proposed approach\label{paInd_MAPE}}
\resizebox{\textwidth}{!}
{\begin{tabular}{|c|cc|cc||cc|cc|}
\toprule
& \multicolumn{4}{c||}{Day ahead forecast} & \multicolumn{4}{c|}{15-min-ahead forecast} \\
\midrule
& \multicolumn{2}{c|}{1 model} & \multicolumn{2}{c||}{fusion} & \multicolumn{2}{c|}{1 model} & \multicolumn{2}{c|}{fusion} \\
\midrule
& MAPE [\%] & score [\%] & MAPE [\%] & score [\%] & MAPE [\%] & score [\%] & MAPE [\%] & score [\%]\\ \midrule
    Ind.1 & 18.0 & 67.4  & 13.8 & 82.8  & 6.2  & 96.1 & 6.1  & 96.2\\
    Ind.2 & 18.4 & 84.1  & 10.4 & 97.9  & 10.7 & 86.3 & 10.2 & 88.2\\
    Ind.3 & 24.4 & 69.2  & 13.6 & 87.4  & 6.6  & 96.5 & 6.5  & 96.4\\
    Ind.4 & 4.3  & 100.0 & 4.6  & 100.0 & 5.4  & 98.6 & 5.2  & 98.8\\
\bottomrule
\end{tabular}}{}
\end{table*}
\begin{table*}[!ht]
\centering
\caption{Comparison of MAE results for the industrial consumers with the proposed approach\label{paInd_MAE}}
\resizebox{\textwidth}{!}
{\begin{tabular}{|c|c|cc|cc||c|cc|cc|}
\toprule
& \multicolumn{5}{c||}{Day ahead forecast} & \multicolumn{5}{c|}{15-min-ahead forecast} \\ \midrule
& & \multicolumn{2}{c|}{1 model} & \multicolumn{2}{c||}{fusion} & & \multicolumn{2}{c|}{1 model} & \multicolumn{2}{c|}{fusion} \\
\midrule
& Thr &MAE [kW] & score [\%] & MAE [kW] & score [\%] &Thr & MAE [kW] & score [\%] & MAE [kW] & score [\%] \\ \midrule
    Ind.1 & 10   & 8.1  & 90.3 & 7.3  & 92.6 & 7.5  & 3.9 & 84.4 & 3.7 & 84.6\\
    Ind.2 & 25   & 12.1 & 72.1 & 9.5  & 86.1 & 18.8 & 7.1 & 92.5 & 6.8 & 92.9\\
    Ind.3 & 22.5 & 19.5 & 54.9 & 16.1 & 82.8 & 16.9 & 8.5 & 86.5 & 8.4 & 86.5\\
    Ind.4 & 15   & 5.9  & 98.7 & 6.2  & 98.7 & 11.3 & 7.6 & 78.1 & 7.4 & 80.8\\
\bottomrule
\end{tabular}}{}
\end{table*}

This approach enables individual models to more effectively learn the intrinsic patterns of load and significantly outperforms other single-model approaches. However, a potential limitation of this approach is that it relies on accurate holiday information, which can compromise flexibility for the sake of precision.

\subsubsection{Commercial consumers}
\begin{figure}[h]
    \centering 
    \includegraphics[width=0.9\textwidth]{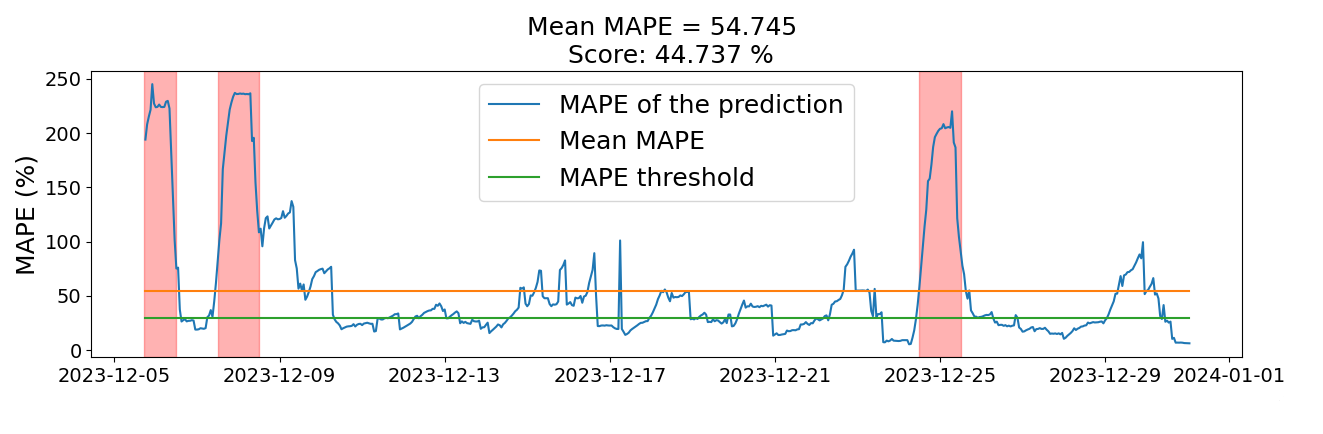}
    \caption{Evolution of MAPE on a commercial consumer}
    \label{fig: MAPE-Com}
\end{figure}

For commercial consumers too, the chosen model is LightGBM, as it is the best performing one according to the results in \ref{sec_append}. Nevertheless, when we again plot the MAPE of the day-ahead forecasters over one month of testing data, in Figure \ref{fig: MAPE-Com}, we observe that the model underperforms for the second commercial customer on dates corresponding to holidays. The marked area highlights two Spanish national holidays on December 6th and 8th, as well as Christmas Day in 2023.

This, however, only happens for this second customer, and in the day-ahead forecaster as the results of the MAPE and MAE and the achieved scores are gathered in Table \ref{paComm_MAPE} and \ref{paComm_MAE} show. Again, the main error source is holidays, so to solve that, again the two-model fusion approach, depicted in Figure \ref{fig: fusion}, has been applied, in the same way as it was done for industrial consumers.

\begin{table}[!ht]
\centering
\caption{Comparison of MAPE results for the commercial consumers with the proposed approach\label{paComm_MAPE}}
\resizebox{\textwidth}{!}
{\begin{tabular}{|c|cc|cc||cc|cc|}
\toprule
& \multicolumn{4}{c||}{Day ahead forecast} & \multicolumn{4}{c|}{15-min-ahead forecast} \\
\midrule
& \multicolumn{2}{c|}{1 model} & \multicolumn{2}{c||}{fusion} & \multicolumn{2}{c|}{1 model} & \multicolumn{2}{c|}{fusion} \\
\midrule
& MAPE [\%] & score [\%] & MAPE [\%] & score [\%] & MAPE [\%] & score [\%] & MAPE [\%] & score [\%]\\ \midrule
    Com.1 & 5.17 & 99.7 & 5.92 & 100  & 6.9  & 90.6 & 7.09 & 91.1 \\
    Com.2 & 48.2 & 70.6 & 26.4 & 87.2 & 15.9 & 64.9 & 17.6 & 61.9 \\
\bottomrule
\end{tabular}}{}
\end{table}
\begin{table}[!ht]
\centering
\caption{Comparison of MAE results for the commercial consumers with the proposed approach\label{paComm_MAE}}
\resizebox{\textwidth}{!}
{\begin{tabular}{|c|c|cc|cc||c|cc|cc|}
\toprule
& \multicolumn{5}{c||}{Day ahead forecast} & \multicolumn{5}{c|}{15-min-ahead forecast} 
\\ \midrule
& & \multicolumn{2}{c|}{1 model} & \multicolumn{2}{c||}{fusion} & & \multicolumn{2}{c|}{1 model} & \multicolumn{2}{c|}{fusion} \\
\midrule
& Thr &MAE [kW] & score [\%] & MAE [kW] & score [\%] & Thr & MAE [kW] & score [\%] & MAE [kW] & score [\%] \\ \midrule
    Com.1 & 6 & 2.9 & 100.0 & 3.3 & 99.7 & 4.5 & 3.9 & 64.1 & 4.05 & 64.6 \\
    Com.2 & 8 & 3.1 & 79.77 & 2.2 & 95.6 & 6   & 1.3 & 95.0 & 1.44 & 95.8 \\
\bottomrule
\end{tabular}}{}
\end{table}

The results indicate that, although the MAPE and MAE values are below the established thresholds in both consumers as well, the quantitative score obtained from a single model does not meet the required standards for the second consumer. In this scenario, the model fusion approach yields better outcomes, reducing both MAPE and MAE while also improving the quantitative score. For the first consumer, despite a slight increase in MAPE and MAE, the values remain well below the thresholds, and the quantitative score overall suggests strong model performance over extended periods.

\subsubsection{Residential consumers}
Forecasting residential power demand presents unique challenges that differ significantly from those encountered in predicting commercial or industrial power demand, as can be perceived from the results gathered in Section \ref{sec_append}. One of the primary difficulties in residential demand forecasting is the high degree of variability and unpredictability in individual consumption patterns. Unlike commercial or industrial settings, where power usage is governed by more consistent schedules and operational needs, residential energy consumption is influenced by a wide range of factors including individual behaviors, household routines, and the use of various home appliances—factors for which we lack detailed information. These factors can lead to sudden and irregular fluctuations in demand, making it harder to model and predict accurately.

\begin{figure}[h]
    \centering 
    \includegraphics[width=0.8\textwidth]{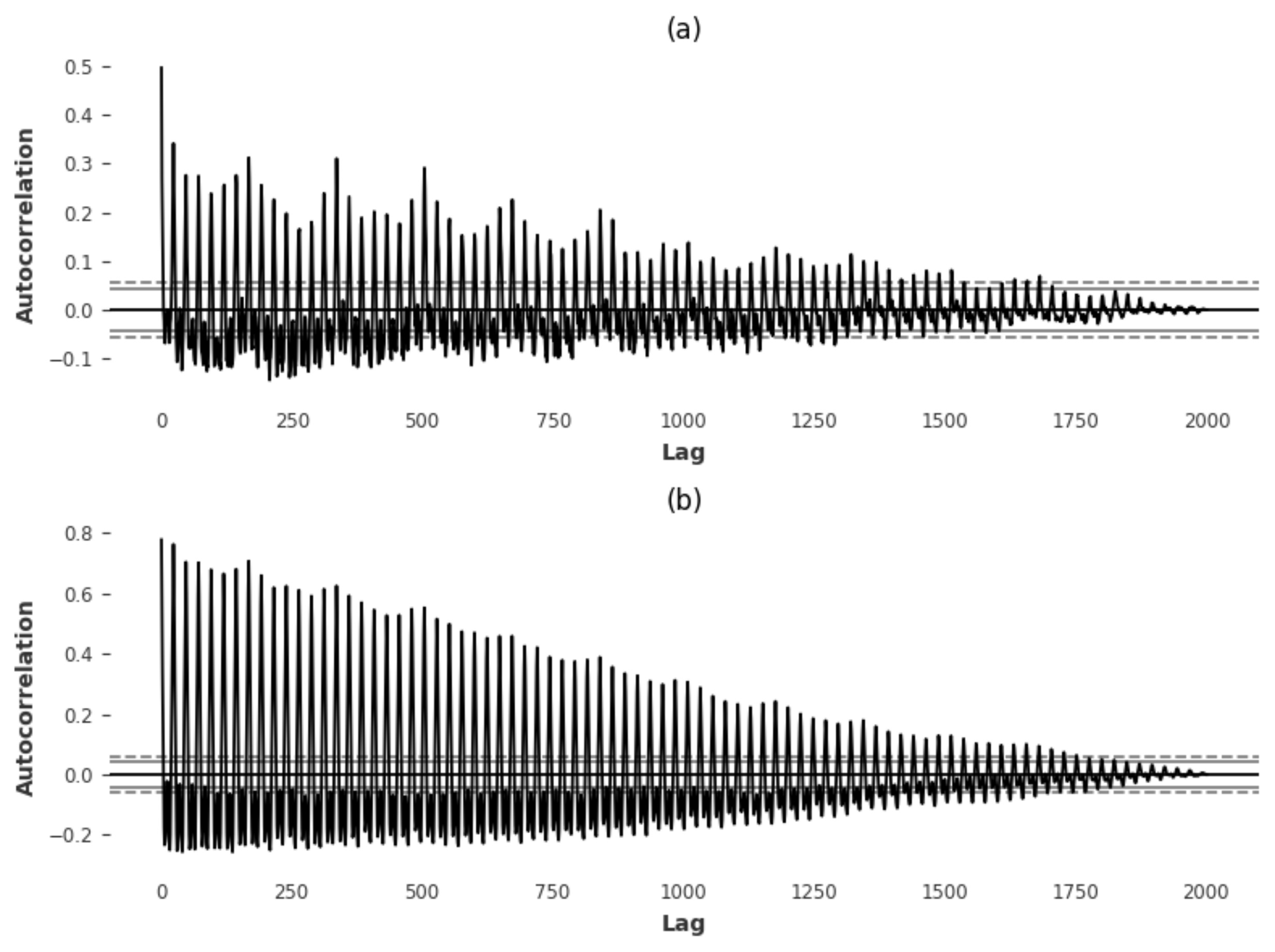}
    \caption{Autocorrelation plot of (a) a municipality with irregular peaks, and (b) a municipality with regular peaks.}
    \label{fig: autocorr}
\end{figure}

Thus, the datasets are in general irregular, with peaks at given hours. Some of the peaks are very periodic and are produced at the same hour every day, but other peaks vary more in time and are more difficult to predict. This fact is described by the weekly and daily autocorrelation plots, in Figure \ref{fig: autocorr}. In some of the municipalities, peaks are more regular and therefore the autocorrelation is considerable, but in other municipalities, autocorrelation is lower, meaning irregularities in its peaks in demand. That is why here again, it is not sufficient to apply a single model directly.

\paragraph{\textbf{Proposed approach}}
In this case, the issue is not related to identifying holidays or the influence of the bimodal behavior of customers, but rather due to the high variability in consumption patterns. This is why this time we seek to reinforce the memory of the model, combining a time series forecasting model with a baseline prediction. The study also considers analogous baseline-refinement approaches, as exemplified in reference \cite{baseline_ref}. 

Adding this baseline prediction offers several advantages in enhancing predictive accuracy and robustness; A time series model captures temporal patterns, trends, and seasonalities inherent in historical data, providing nuanced and dynamic forecasts. However, baseline predictions, often simpler and based on historical averages or other straightforward methods, serve as a reliable reference point. By integrating these two approaches, the model benefits from the sophisticated pattern recognition of time series analysis while maintaining the stability and simplicity of baseline predictions. This combination helps mitigate the risk of overfitting, improves performance during anomalous events or data fluctuations such as the mentioned irregularities in the peak demand, and ensures more consistent and reliable predictions. 

Regarding the day-ahead forecast, for every point, the baseline method calculates the arithmetic mean of the consumption on the last day and the last week at the same hour, as in equation (5). Then, the series of the baseline prediction is used as an exogenous variable in a gradient-boosting model.

\begin{equation}
    x_i = 0.5\cdot(x_{i-24} + x_{i-7\cdot24})
\end{equation}

For the 15-minute-ahead forecast, the baseline has been calculated following the equation (6). The assigned weights are 0.6 to the last point, 0.2 to the mean consumption of the last 4 days at that same time, and 0.2 to the mean consumption of the last 4 weeks at the same time and weekday. This way, the baseline has a memory of the consumption during the last days, not only the last value.

\begin{multline}
x_i = 0.6 \cdot \bar{x}_{i-1} + \frac{0.2}{4}\ \cdot (x_{i-24} + x_{i-48} + x_{i-72} + x_{i-96}) + \frac{0.2}{4}\ \cdot (x_{i-24\cdot7} + x_{i-48\cdot7} + x_{i-72\cdot7} + x_{i-96\cdot7})
\end{multline}

In general, the results gathered in Section \ref{sec_append} suggest that the predictability of these consumers, with highly variable patterns, is further challenged by the relatively low magnitude of residential consumption. The model prioritizes accuracy in predicting the lowest consumption values because errors at low levels significantly increase the loss function in the case of MAPE, but then the peak demand predictions are missed. Conversely, prioritizing the accuracy in peak demand points highly increases the MAPE, as the error increases in the low consumption part. 

Furthermore, as shown in equation (4), in order to calculate the MAPE, it is necessary to divide by the actual value of the dataset. For the \textit{residential} consumers, these values are often less than 1 kW, which leads to significantly higher MAPE values due to the division by such small numbers. That is why, for the case of \textit{residential} consumers, the thresholds of the MAPE are raised up to 30\% for the day-ahead forecasts and to 25\% for the 15-minute-ahead forecasts. The results comparing one model with the proposed hybrid approach are shown in Tables \ref{paRes_MAPE} and \ref{paRes_MAE}.

\begin{table}[t]
\centering
\caption{Comparison of MAPE results for the residential consumers with the proposed approach\label{paRes_MAPE}}
\resizebox{\textwidth}{!}
{\begin{tabular}{|c|c|cc|cc||c|cc|cc|}
\toprule
& \multicolumn{4}{c||}{Day ahead forecast} & \multicolumn{4}{c|}{15-min-ahead forecast} \\
\midrule
& \multicolumn{2}{c|}{1 model} & \multicolumn{2}{c||}{hybrid} & \multicolumn{2}{c|}{1 model} & \multicolumn{2}{c|}{hybrid} \\
\midrule
& MAPE [\%] & score [\%] & MAPE [\%] & score [\%] & MAPE [\%] & score [\%] & MAPE [\%] & score [\%]\\ \midrule
    Res.1 & 22.5 & 96.4 & 21.9 & 95.4 & 10.4 & 90.0 & 9.2 & 95.1 \\
    Res.2 & 28.6 & 67.9 & 28.6 & 68.4 & 10.5 & 92.8 & 9.9 & 93.5 \\
    Res.3 & 21.6 & 88.6 & 20.8 & 91.8 & 11.2 & 88.7 & 10.2 & 92.8\\
    Res.4 & 21.2 & 97.4 & 20.6 & 98.0 & 12.5 & 88.6 & 11.7 & 91.2\\
    Res.5 & 31.7 & 42.1 & 30.3 & 47.9 & 9.3  & 94.4 & 9.1 & 95.4\\
\bottomrule
\end{tabular}}{}
\end{table}
\begin{table}[!ht]
\centering
\caption{Comparison of MAE results for the residential consumers with the proposed approach\label{paRes_MAE}}
\resizebox{\textwidth}{!}
{\begin{tabular}{|c|c|cc|cc||c|cc|cc|}
\toprule
& \multicolumn{5}{c||}{Day ahead forecast} & \multicolumn{5}{c|}{15-min-ahead forecast} \\ \midrule
& & \multicolumn{2}{c|}{1 model} & \multicolumn{2}{c||}{hybrid} & & \multicolumn{2}{c|}{1 model} & \multicolumn{2}{c|}{hybrid} \\
\midrule
& Thr &MAE [kW] & score [\%] & MAE [kW] & score [\%] & Thr & MAE [kW] & score [\%] & MAE [kW] & score [\%] \\ \midrule
    Res.1 & 0.61 & 0.53 & 88.2 & 0.53 & 83.4 & 0.46 & 0.06 & 99.8 & 0.05 & 99.8\\
    Res.2 & 0.79 & 0.78 & 62.5 & 0.78 & 63.5 & 0.58 & 0.07 & 99.9 & 0.07 & 99.9\\
    Res.3 & 1.20 & 0.87 & 93.4 & 0.87 & 93.1 & 0.90 & 0.11 & 99.9 & 0.10 & 100 \\
    Res.4 & 0.39 & 0.39 & 57.2 & 0.38 & 61.8 & 0.29 & 0.06 & 98.9 & 0.05 & 99.0\\
    Res.5 & 0.43 & 0.56 & 15.6 & 0.55 & 19.9 & 0.32 & 0.04 & 99.4 & 0.04 & 99.5\\
\bottomrule
\end{tabular}}{}
\end{table}

The results indicate that, although the MAPE and MAE values are below the established thresholds, or very close to them, even when applying a single model, the quantitative score obtained does not meet the required standards for consumers 2 and 5 in the day-ahead forecast, suggesting that the model lacks robustness. 

The hybrid approach yields improved results overall, reducing both MAPE and MAE while also increasing the quantitative score values, although not enough to meet the quantitative score thresholds for consumers 2 and 5. These modest improvements underscore the complexity of forecasting residential demand. However, the model's overall performance is enhanced over extended periods, demonstrating its superior effectiveness and stability.

\subsection{Discussion of results}

The findings of this study are consistent with those of previous research, underscoring the significance of consumer-specific forecasting models. The classification algorithm employed in this study attained accuracies that are analogous to those reported in studies such as \cite{Toit2016} and \cite{Cluster3}. These studies emphasised the advantages of customer segmentation, though we acknowledge the need for additional consumers to be included as a future work to ensure more extensive validation. For industrial consumers, calendar-related features and holidays were key predictors, supported by findings in \cite{ML2}. The fusion approach improved accuracy during holidays, addressing challenges noted in \cite{XGB}.
Commercial consumers showed similar dependencies on calendar features, with weather conditions, particularly temperature, playing a significant role, as seen in \cite{Perera2020} and \cite{Luo2021}. Residential forecasting proved most challenging due to high variability, consistent with \cite{Yu2023} and \cite{baseline_ref}. Our hybrid approach, combining time series forecasting with baseline predictions, improved robustness but highlighted the complexity of residential demand.
Overall, this study reinforces the value of tailored forecasting strategies and advanced machine learning techniques, as seen in \cite{ML1} and \cite{feature4}, offering DSOs tools to enhance grid stability and efficiency.

\section{Conclusions}
In conclusion, this study has achieved several milestones in the development and application of forecasting methodologies for different consumer types. We have successfully created a straightforward yet highly accurate customer classification system that effectively differentiates between industrial, commercial, and residential consumers. This classification system is foundational for tailoring forecasting models to the specific needs of each consumer type.

Our feature analysis has provided critical insights into the most significant variables influencing each consumer category. For industrial consumers, calendar-related features and holiday information emerged as the primary drivers of demand, with weather conditions proving to have negligible effects. In the case of commercial consumers, while calendar-related features and holidays remain crucial, weather conditions also play a notable role in influencing consumption patterns. Residential consumers, on the other hand, exhibit a more nuanced interplay of factors, where calendar-related features, holiday information, and weather variables all contribute, albeit to a lesser extent.

Furthermore, we explored a diverse array of AI and machine learning algorithms for load forecasting, rigorously comparing their performance to identify the most effective methods. When initial forecasting results did not meet the required standards, we introduced novel forecasting approaches that demonstrated superior performance compared to the previously tested models. In addition, it is to be mentioned that we developed and tested the forecasters as if they were deployed in an online environment; in the case of the day-ahead forecaster, a production scenario was simulated by predicting the next 24 hours at each time step of the testing data, rather than making a single prediction every 24 hours. This approach more accurately reflects real-world conditions and enhances the reliability of the models.

In conclusion, it is imperative to emphasise that the primary objective of the study was to formulate a comprehensive methodology for the classification and forecasting of load across diverse consumer types. The efficacy of grid planning and operation is contingent on the accuracy of load forecasts. Consequently, the study sought to ascertain the impact of distinct characteristics and forecasting models on STLF and VSTLF, respectively, for each consumer category. The insights derived from this research are designed to serve as a guide for DSOs, aiming to enhance their grid planning, synchronisation, and operational capabilities.

\section{Future work}

In this section, we outline potential directions for advancing the research presented in this study. The primary goal is to deploy the developed models in an online environment, enabling real-time demand forecasting. This deployment involves several critical steps to ensure the system's effectiveness and reliability. First, the input data must undergo thorough preprocessing, including detecting and addressing outliers, as well as performing data imputation to fill any gaps or inconsistencies that could compromise the accuracy of the forecasts.

Once the input data is properly processed, the classification algorithms and forecasting models would be deployed in an online environment allowing the models to continuously receive and process new data, generating up-to-date predictions. To maintain and enhance the accuracy and relevance of the models over time, it is crucial to implement iterative refinement processes. These could include Change Point Detection, which identifies significant shifts in data patterns that could affect model performance; Concept Drift Detection, which monitors and adapts to changes in the underlying data distribution; and Model Retraining, which periodically updates the models based on the latest available data. By incorporating these refinement processes, the forecasting models will be better equipped to sustain high performance and adapt to evolving conditions, ensuring their long-term utility in managing the electrical grid effectively.

\section{Acknowledgements}
This research was funded by the European Union’s Horizon Europe research and innovation programme under HORIZON-EUSPA-2021-SPACE call, grant agreement No. 101082355, with the acronym RESPONDENT.

\newpage

\appendix
\section{}\label{sec_append}

\begin{table}[!ht]
\centering
\caption{Error metrics of 24 hours ahead models
\label{tab_results_24}}
{\begin{tabular}{|c|cccc|cc|ccccc|}
\toprule
Model &Ind.1 &Ind.2 &Ind.3 &Ind.4 &Com.1 &Com.2 &Res.1 &Res.2 &Res.3 &Res.4 &Res.5 \\
\midrule
MAPE (\%) &&&&&&&&&&&\\
\midrule
Baseline &64.31 &60.77 &63.03 &8.70 &14.97 &87.05 &31.32 &36.77 &24.96 &28.15 &41.81\\
MLP      &40.39   &30.39   &30.51   &5.18 &7.75  &80.02  &33.42 &36.07 &24.01 &27.52 &36.39 \\
LSTM     &22.28   &\textbf{13.45}   &\textbf{17.95}   &4.58  &5.77 &\textbf{26.80} &33.77 &35.15 &26.57 &30.14 &33.79 \\
RF       &41.66   &29.46   &39.86   &6.39 &7.28 &59.77 &27.15 &32.69 &27.23 &22.34 &33.66 \\
SVR      &38.96   &27.18   &32.75   &6.71 &6.87 &78.97 &24.00 &29.27 &\textbf{20.18} &21.54 &\textbf{30.68} \\
XGBoost  &26.17   &17.13   &29.43   &4.62 &5.76 &44.91 &23.06 &29.20 &23.99 &22.27 &32.85 \\
LightGBM &\textbf{18.01} &18.39 &24.37 &\textbf{4.33} &\textbf{5.17} &48.17 &\textbf{22.52} &\textbf{28.56} &21.64 &\textbf{21.19} &31.67 \\
\midrule
MAE (kW/h) &&&&&&&&&&&\\
\midrule
Baseline &22.16 &34.50 &55.76 &12.34 &8.19 &4.84 &0.76 &1.03 &1.04 &0.52 &0.73 \\
MLP      &12.67 &16.43 &22.95 &6.84 &4.36 &3.28 &0.73 &0.90 &0.94 &0.50 &0.63 \\
LSTM     &10.24 &\textbf{10.49} &\textbf{17.36} &6.19 &3.36 &\textbf{1.52} &0.69 &0.86 &1.12 &0.48 &0.58 \\
RF       &14.37 &21.29 &27.84 &8.86 &4.15 &2.81 &0.61 &0.84 &1.04 &0.40 &0.58 \\
SVR      &11.90 &17.09 &23.78 &8.77 &3.97 &3.72 &0.59 &0.85 &0.91 &0.41 &0.60 \\
XGBoost  &9.51 &12.18 &21.21 &6.29 &3.23 &3.07 &0.55 &0.79 &0.95 &0.40 &0.58 \\
LightGBM &\textbf{8.13} &12.09 &19.47 &\textbf{5.95} &\textbf{2.95} &3.14 &\textbf{0.53} &\textbf{0.78} &\textbf{0.87} &\textbf{0.39} &\textbf{0.56} \\
\bottomrule
\end{tabular}}{}
\end{table}
\begin{table}[!ht]
\centering
\caption{Error metrics of 15 minutes ahead models
\label{tab_results_15}}
{\begin{tabular}{|c|cccc|cc|ccccc|}
\toprule
Model &Ind.1 &Ind.2 &Ind.3 &Ind.4 &Com.1 &Com.2 &Res.1 &Res.2 &Res.3 &Res.4 &Res.5 \\
\midrule
MAPE (\%) &&&&&&&&&&&\\
\midrule
Baseline &6.90 &14.13 &7.24 &7.02 &10.05 &21.88 &12.19 &11.99 &10.16 &14.81 &11.26 \\
MLP      &7.72 &12.45 &6.79 &5.77 &7.44 &27.90 &9.54 &10.09 &9.87 &12.40 &9.39 \\
LSTM     &9.79 &13.45 &7.66 &6.28 &9.02 &26.79 &9.84 &10.65 &10.74 &12.27 &9.54 \\
RF       &7.19 &11.02 &6.90 &5.76 &7.87 &24.18 &9.24 &\textbf{9.79} &9.28 &\textbf{11.98} &\textbf{9.26} \\
SVR      &9.63 &13.77 &6.93 &5.58 &7.62 &25.88 &\textbf{9.17} &9.82 &\textbf{9.02} &12.36 &9.45 \\
XGBoost  &\textbf{6.45} &10.88 &\textbf{6.54} &5.50 &7.45 &21.17 &9.33 &10.27 &10.86 &12.09 &9.37 \\
LightGBM &6.58 &\textbf{10.63} &6.65 &\textbf{5.46} &\textbf{6.94} &\textbf{15.93} &10.44 &10.47 &11.23 &12.49 &9.28 \\
\midrule
MAE (kW/h) &&&&&&&&&&&\\
\midrule
Baseline &4.58 &8.99 &9.74 &9.88 &5.86 &1.85 &0.074 &0.083 &0.109 &0.066 &0.052 \\
MLP      &4.49 &7.92 &8.90 &8.11 &4.25 &1.63 &0.058 &0.068 &0.101 &0.055 &0.043 \\
LSTM     &4.98 &9.82 &7.99 &8.72 &5.18 &1.66 &0.059 &0.070 &0.108 &\textbf{0.053} &0.043 \\
RF       &4.63 &7.51 &9.16 &8.15 &5.50 &1.63 &0.057 &\textbf{0.067} &0.098 &0.054 &\textbf{0.042} \\
SVR      &4.57 &8.25 &8.64 &7.85 &4.42 &1.63 &\textbf{0.056} &\textbf{0.067} &\textbf{0.097} &0.055 &0.043 \\
XGBoost  &4.08 &7.71 &8.62 &7.78 &4.28 &1.62 &\textbf{0.056} &0.069 &0.108 &0.054 &0.043 \\
LightGBM &\textbf{3.92} &\textbf{7.01} &\textbf{8.27} &\textbf{7.73} &\textbf{3.96} &\textbf{1.33} &0.060 &0.069 &0.109 &0.055 &\textbf{0.042} \\
\bottomrule
\end{tabular}}{}
\end{table}

\newpage

\bibliographystyle{elsarticle-num} 
\bibliography{cas-refs.bib}

\end{document}